\documentclass[10pt,twocolumn,letterpaper]{article}

\usepackage{cvpr}
\usepackage{times}
\usepackage{epsfig}
\usepackage{graphicx}
\usepackage{amsmath}
\usepackage{amssymb}
\usepackage{pbox}

\usepackage[table]{xcolor}
\definecolor{bedColor}{rgb}{0, 0, 1}
\definecolor{booksColor}{rgb}{0.9137,0.3490,0.1882}
\definecolor{ceilColor}{rgb}{0, 0.8549, 0}
\definecolor{chairColor}{rgb}{0.5843,0,0.9412}
\definecolor{floorColor}{rgb}{0.8706,0.9451,0.0941}
\definecolor{furnColor}{rgb}{1.0000,0.8078,0.8078}
\definecolor{objsColor}{rgb}{0,0.8784,0.8980}
\definecolor{paintColor}{rgb}{0.4157,0.5333,0.8000}
\definecolor{sofaColor}{rgb}{0.4588,0.1137,0.1608}
\definecolor{tableColor}{rgb}{0.9412,0.1373,0.9216}
\definecolor{tvColor}{rgb}{0,0.6549,0.6118}
\definecolor{wallColor}{rgb}{0.9765,0.5451,0}
\definecolor{windColor}{rgb}{0.8824,0.8980,0.7608}

\usepackage{listings}

\usepackage[breaklinks=true,bookmarks=false]{hyperref}
\hypersetup{colorlinks=true,urlcolor=red,linkcolor=red,citecolor=green}

\cvprfinalcopy % *** Uncomment this line for the final submission

\begin{document}

%%%%%%%%% TITLE
\title{SceneNet: Understanding Real World Indoor Scenes With Synthetic Data}

\author{Ankur Handa,
%\and
Viorica P{\u a}tr{\u a}ucean,
%\and
Vijay Badrinarayanan,
%\and
Simon Stent and
\and
\and
Roberto Cipolla\\
\\
Department of Engineering,\\
University of Cambridge \\
{\tt\small \{ah781,vp344,vb292,sais2,rc10001\}@cam.ac.uk}
}

\maketitle
%\thispagestyle{empty}

%%%%%%%%% ABSTRACT
\begin{abstract}
   Scene understanding is a prerequisite to many high level tasks for any automated intelligent machine operating in real world environments. Recent attempts with supervised learning have shown promise in this direction but also highlighted the need for enormous quantity of supervised data --- performance increases in proportion to the amount of data used. However, this quickly becomes prohibitive when considering the manual labour needed to collect such data. In this work, we focus our attention on depth based semantic per-pixel labelling as a scene understanding problem and show the potential of computer graphics to generate virtually unlimited labelled data from synthetic 3D scenes. By carefully synthesizing training data with appropriate noise models we show comparable performance to state-of-the-art RGBD systems on NYUv2 dataset despite using only depth data as input and set a benchmark on depth-based segmentation on SUN RGB-D dataset. Additionally, we offer a route to generating synthesized frame or video data, and understanding of different factors influencing performance gains.
\end{abstract}

%%%%%%%%% BODY TEXT
\section{Introduction}
Many high level tasks in real world require some knowledge of objects present in the scene, their physical locations and the underlying physics involved as a means to understanding the scene. Autonomously navigating robots equipped with cameras can build upon the awareness and understanding of their environment to perform a range of simple to complex tasks in real world. In this work, we choose to focus on per-pixel semantic labelling as our scene understanding problem. We believe semantic understanding can provide important and rich information that is needed to carry out many high level tasks. In a broader sense, it gives an estimate of volume and actual physical extent of the objects \textit{e.g.} an indoor robot must understand how the floor bends or curves so as to adjust its path while navigating, a robot operating in a kitchen may need to have an idea of the volume of the objects to appropriately arrange them in the cabinet --- knowledge of the physical extent of the supporting surface can provide a rough estimate of where different objects can be placed. In 3D modelling, it is sometimes required to have precise knowledge of where different objects can be fitted or inserted into others. It is exactly in these high level tasks where the role of semantic segmentation is emphasized more than pure object detection. Furthermore, it is more likely to benefit from the context present in the scene to segment objects unlike any `blanket' detection module that is generally run independently on a sliding window on the image.
 
Motivated by its recent success, we use deep learning as our computational framework for semantic segmentation. Modelled ostensibly on the human brain \cite{Cadieu:etal:PLOS2014}, deep learning models have superseded many traditional approaches that relied on hand engineered features --- past few years have seen a rapid proliferation of deep learning based approaches in many domains in AI. However, a major limitation of modern day deep learning models is the requirement of large quantity of supervised training data. Collecting big datasets can quickly become labour intensive and may not be a viable option. In this work, we focus on the challenges of obtaining the desired training data for scene understanding.

Many existing datasets do not have the volume of data needed to make significant advances in scene understanding that we aim in this work. For instance, considering indoor scene datasets which we focus on, NYUv2 \cite{Silberman:etal:ECCV2012} contains only 795 training images for as many as 894 object classes. SUN RGB-D \cite{Song:etal:CVPR2015}, on the other hand contains 5,285 training images for 37 classes. These are the only two indoor depth datasets with per-pixel labels and are limited in size considering the enormity of data needed to achieve good performance on unseen data. Both relied on humans for labelling which can quickly become tedious and expensive process. We believe scene understanding can greatly benefit from the computer graphics community that has long had the tradition of CAD model repositories. Synthetic data is already used for many computer vision problems \cite{Fischer:etal:arXiv2015,Aubry:etal:CVPR2014,Aubry:etal:ICCV2015,Gupta:etal:ECCV2014,Gupta:etal:CVPR2015,Jiajun:etal:NIPS2015} and \cite{Handa:etal:ICRA2014,Handa:etal:ECCV2012} in the context of robotics. We believe that the role of synthetic data and gaming environments \cite{Mnih:etal:Nature2015} will continue to grow in providing training data with further advances in machine learning and data driven understanding. 

Our main contribution in this work is to propose a new dataset of annotated 3D scenes which can generate virtually unlimited ground truth training data and show its potential in improving the performance of per-pixel labelling on challenging real world indoor datasets.

\section{Related Work}
\label{sec:relatedwork}
\begin{figure}%[htp]
\centerline{
\includegraphics[width=0.25\linewidth]{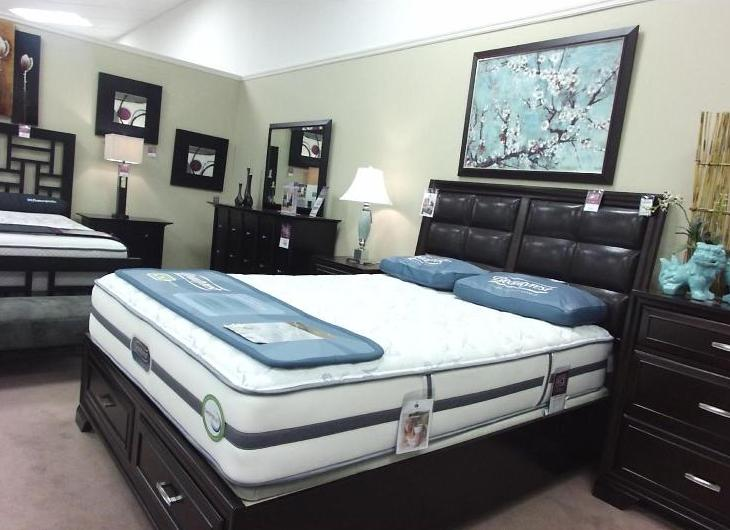}
\includegraphics[width=0.25\linewidth]{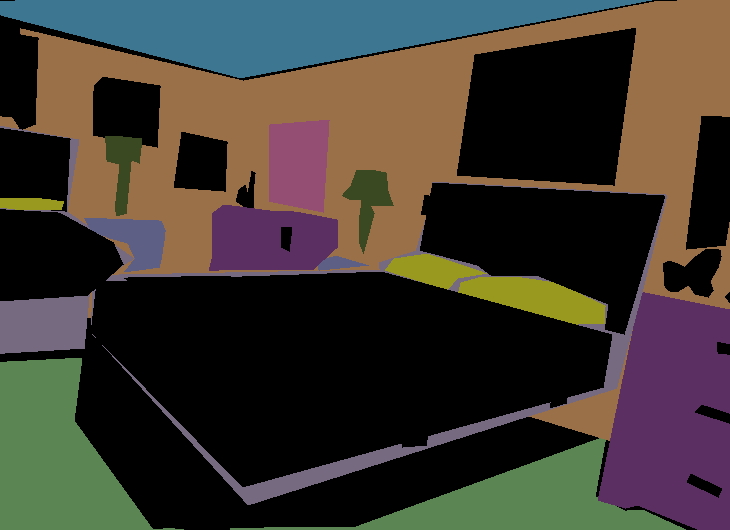}
\includegraphics[width=0.25\linewidth]{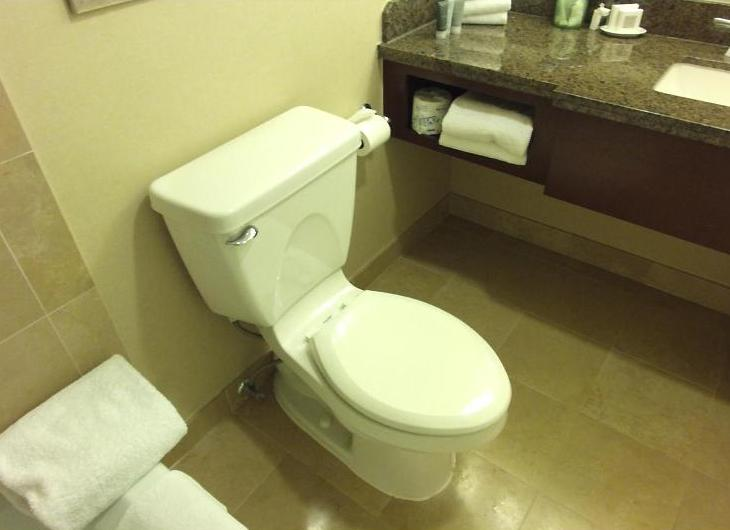}
\includegraphics[width=0.25\linewidth]{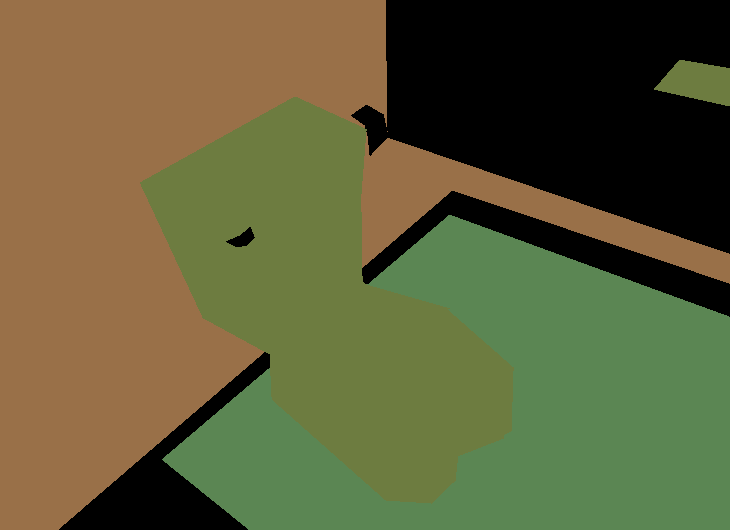}
}
\centerline{
\hfill
\makebox[0.50\linewidth][c] { \footnotesize{(a) Missing labels }}
\hfill
\makebox[0.50\linewidth][c] { \footnotesize{(b) Mislabelled}}
}
\caption{\small
Missing labels (a) and mislabelled frames (b) are very common in many real datasets. In (b) the toilet and sink have the same ground truth label. Both images are from SUN RGB-D\cite{Song:etal:CVPR2015}.} \vspace{1mm}
\label{fig:sunrgbd_wronglabels}
\end{figure}

Prior work on per-pixel indoor image labelling has been due to NYUv2 and SUN RGB-D datasets. The work of \cite{Couprie:etal:ICLR2013} was one of the first in the direction and built on top of a deep learning framework trained on NYUv2. However, it achieved only modest performance on the test data. Subsequently, \cite{Long:etal:CVPR2015} and \cite{Eigen:etal:ICCV2015} have improved the performance, again with deep learning inspired methods. We think that the potential of these new methods is yet to be explored fully and that the lack of training data is the primary hindrance. Both NYUv2 and SUN RGB-D are limited in their sizes and only provide per-pixel labels for low quality raw depth-maps and corresponding RGB frames --- missing data and noise in raw sensor measurements exacerbate the problem even more. Also, since most of the labelling relies on human labour, missing labels and mislabelled data are very common, as shown in Fig. \ref{fig:sunrgbd_wronglabels}. This is inevitable as labelling for humans can be a tiring process and sometimes comes with a considerable monetary cost. 

Ideally, one would like to have a fully labelled 3D model for every scene to generate annotations from arbitrary viewpoints but this is clearly missing in both the datasets. SUN3D \cite{Xiao:etal:ICCV2013} goes in the direction of providing annotated video sequences together with 3D point clouds obtained with SfM. However, they only provide 8 such sequences. The NYUv2 dataset provides a large number of videos, but only provide one annotated frame per video. Additionally, the videos are captured without 3D reconstruction in mind and therefore not suitable for generating accurate 3D reconstructions or annotations, as observed from our own experiments. Furthermore, fusion of raw depth images within a temporal window can provide smooth depth-map to aid the segmentation as opposed to noisy raw depth-maps. \cite{Xiao:etal:ICCV2013} fuse the raw depth maps but again they are limited by the overall variety of annotated sequences. On the other hand, NYUv2 and SUNRGBD do not provide fused depth measurements. Fortunately, both datasets provide an inpainted version of the raw depth maps which has been used in \cite{Couprie:etal:ICLR2013,Long:etal:CVPR2015,Eigen:etal:ICCV2015} on per-pixel scene understanding. In this work, we provide a new way of generating potentially unlimited labelled training data with perfect ground truth inspired from computer graphics. We build our own new library of synthetic indoor scenes, called SceneNet and generate data to train our deep learning algorithm for per-pixel labelling.

The idea of using synthetic scenes has existed in the past, in particular, \cite{Fisher:etal:SIGGRAPHASIA2012}, who released a number of scenes targetted towards the application of scene retrieval, which could potentially be used in the problem we are interested in. However, those scenes are small scale of the order of 4m$\times$3m$\times$3m, and contain only one or two instances of characteristic objects that define the scene \textit{e.g.} only one desk and monitor in the office room. On the other hand, object repositories have existed for a long time now particularly the famous Google Warehouse \cite{Trimble} and now ModelNet \cite{Wu:etal:CVPR2015} and ShapeNet \cite{shapeNet}. Unfortunately, object repositories are not directly useful for the problem we are targetting. Therefore, we compile a number of basis scenes downloaded from the internet and synthesize virtually unlimited number of new scenes by placing objects sampled from these object repositories with various constraints on their joint placement, using simulated annealing \cite{Craigyu:etal:ToG2011}. In doing so, we not only add variety to our dataset we also obtain free annotations automatically thus completely removing human labellers (or MTurk) from the loop.

\begin{figure}[t]
\centerline{
\includegraphics[width=\linewidth]{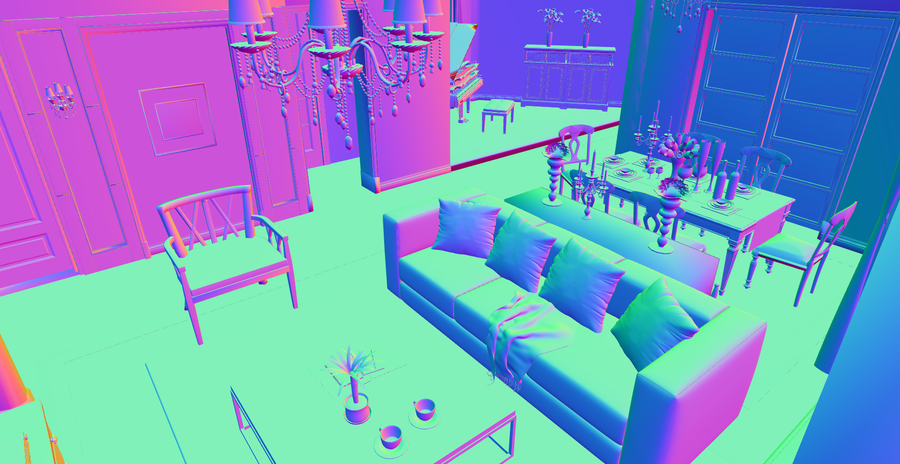}
}
\centerline{
\hfill
\makebox[\linewidth][c] { \footnotesize{(a) Sample basis scene from SceneNet.} }
}
\vspace{1mm}
\centerline{
\includegraphics[width=0.24\linewidth]{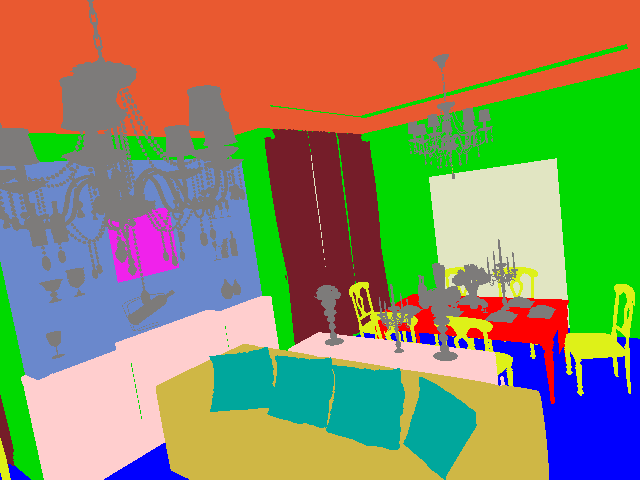}
\includegraphics[width=0.24\linewidth]{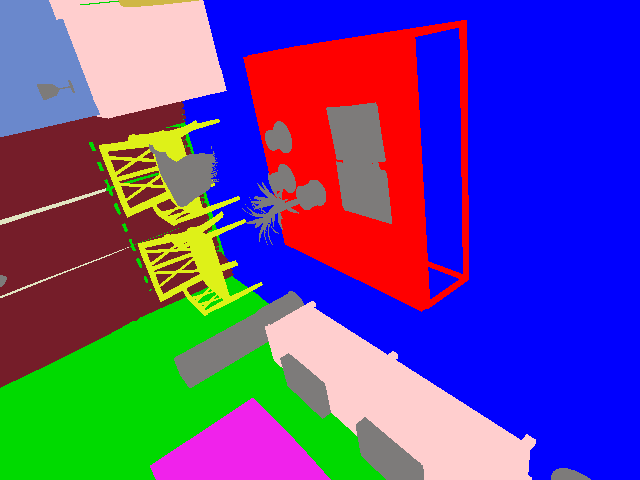}
\includegraphics[width=0.24\linewidth]{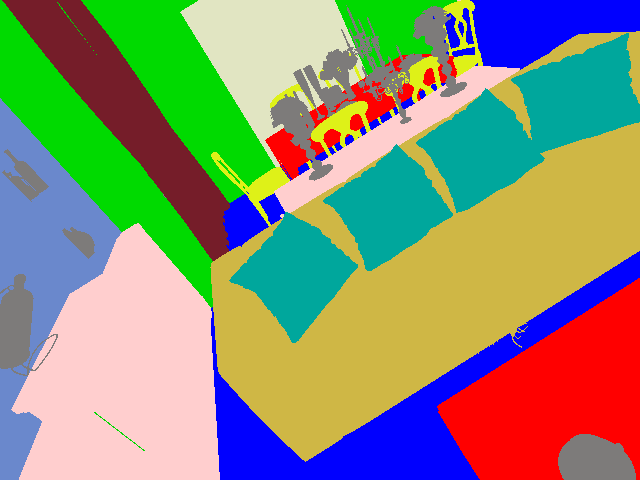}
\includegraphics[width=0.24\linewidth]{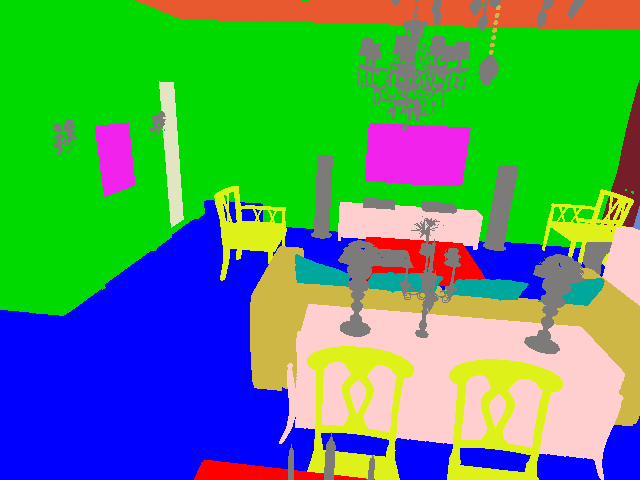}
}
\centerline{
\hfill
\makebox[\linewidth][c] { \footnotesize{ (b) Examples of per-pixel semantically labelled views from the scene.}}
}
\caption{\small
Annotated 3D models allow the generation of per-pixel semantically labelled images from arbitrary viewpoints, such as from a floor-based robot or a UAV. Just as the ImageNet \cite{Deng:etal:CVPR2009} and ModelNet \cite{Wu:etal:CVPR2015} datasets have fostered recent advances in image classification \cite{Krizhevsky:etal:NIPS2012} and 3D shape recognition \cite{Su:etal:ICCV2015}, we propose SceneNet as a valuable dataset towards the goal of indoor scene understanding.} \vspace{1mm}
\label{fig:gt_annotation}
\end{figure}

\section{Synthesizing Training Data}
SceneNet is inspired by the efforts developed in the graphics community to build large scale repositories of CAD models. Since our goal is labelling the scenes, we first build an open-source repository of annotated synthetic indoor scenes 
--- the SceneNet Basis Scenes (SN-BS) --- containing a significant number of manually labelled 3D models. Having a labelled scene gives a lot more flexibility in obtaining the desired training data --- annotations from arbitrary view points are easily available, saving the expensive human labour required to label each image independently. Most importantly, we can also render large numbers of videos from these scenes. This is certainly something that is missing in all the datasets and we foresee that it will play an important role in understanding RGBD videos. However, since we quantify performance on only static images in this work, we generate synthetic data from random poses.
\begin{table}[t]
\centering
\begin{tabular}{|l|c|r|}
\hline
\textbf{Category}      & \textbf{\# 3D models} & \textbf{\# objects}\\ \hline
Bedrooms      & 11       &    428       \\ \hline
Office Scenes & 15       &    1,203       \\ \hline
Kitchens      & 11       &    797       \\ \hline
Living Rooms  & 10       &    715       \\ \hline
Bathrooms     & 10       &    556       \\ \hline
\end{tabular}
\vspace{1mm}
\caption{\small Different scene categories and the number of annotated 3D models 
for each category in SN-BS.}\label{table: SceneNetData}
\end{table}

\begin{figure*}[htp]
\centerline{
\includegraphics[width=0.20\linewidth]{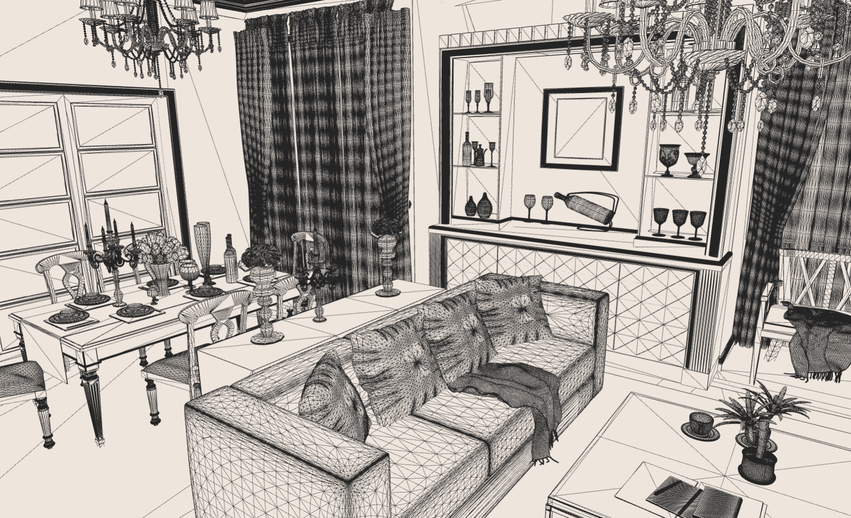}
\includegraphics[width=0.21\linewidth]{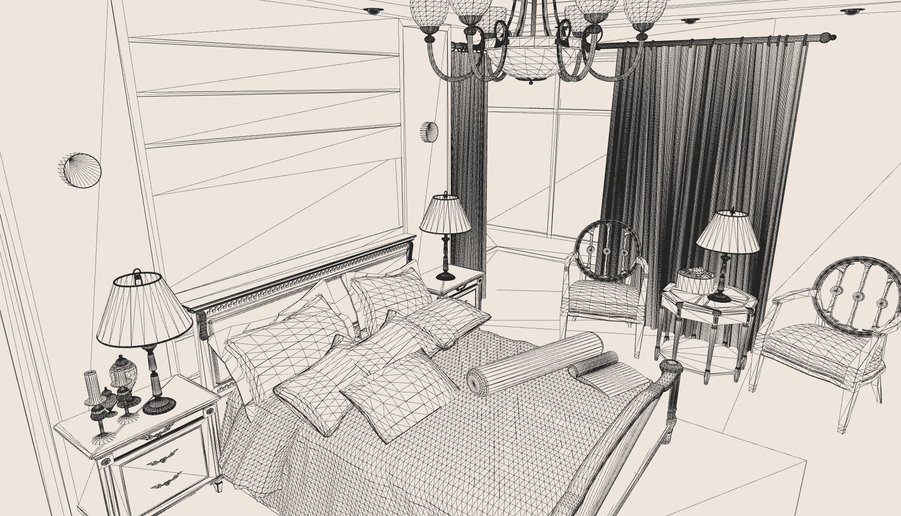}
\includegraphics[width=0.189\linewidth]{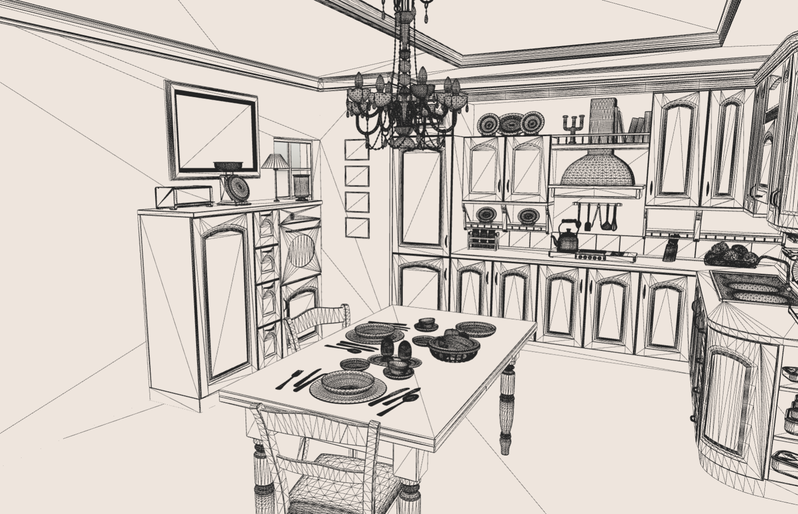}
\includegraphics[width=0.20\linewidth]{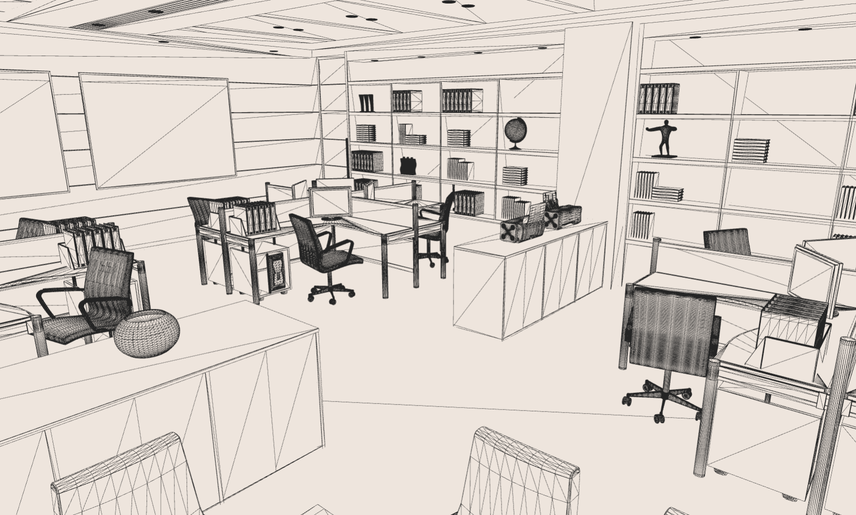}
\includegraphics[width=0.20\linewidth]{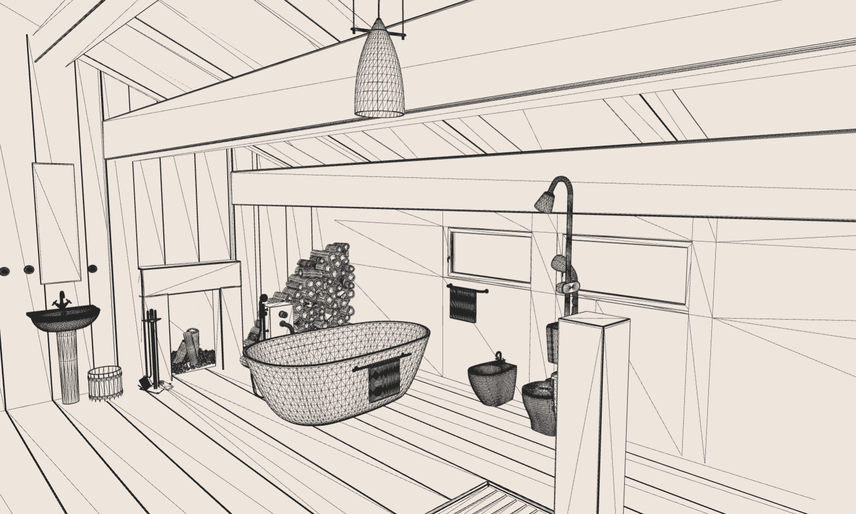}
}
\centerline{
\hfill
\makebox[0.20\linewidth][c] { \footnotesize{(a) Living room }}
\hfill
\makebox[0.21\linewidth][c] { \footnotesize{(b) Bedroom } }
\hfill
\makebox[0.189\linewidth][c] { \footnotesize{(c) Kitchen}}
\hfill
\makebox[0.20\linewidth][c] { \footnotesize{(d) Office}}
\hfill
\makebox[0.20\linewidth][c] { \footnotesize{(e) Bathroom}}
}
\caption{\small
Snapshots of detailed scenes for each category in SceneNet, hosted at \url{robotvault.bitbucket.org}} \vspace{1mm}
\label{fig:SceneNet}
\end{figure*}

SN-BS contains 3D models from five different scene categories illustrated in Fig.~\ref{fig:SceneNet}, with at least 10 annotated scenes 
per category, and have been compiled together from various online 3D repositories \textit{e.g.} \url{www.crazy3dfree.com} and \url{www.3dmodelfree.com}, and wherever needed, manually annotated. Importantly, all the 3D models are metrically accurate. Each scene is 
composed of 15--250 objects (see Table \ref{table: SceneNetData}) but the complexity can be controlled algorithmically. The granularity of the annotations can easily be adapted by the user depending on the application. All the models are available in standard \textit{.obj} format. A simple OpenGL based GUI allows the user to place virtual cameras in the synthetic scene at desired locations to generate a 
possible trajectory for rendering at different viewpoints. Fig. 
\ref{fig:gt_annotation}(b) shows samples of rendered annotated views of a living room. Since the annotations are directly in 3D, objects can be replaced at their respective locations with similar objects sampled from existing 3D object databases to generate variations of the same scene with larger intra-class shape variation. Moreover, objects can be perturbed from their positions and new objects added to generate a wide variety of new scenes. 
We also tried texturing the scenes using freely available OpenSurfaces \cite{Bell:etal:SIGGRAPH2013} but found that the scenes did not reflect a faithful representation of the real world. Raytracing is another option we considered but found it to be very time consuming when modelling complicated light transport. Therefore, we have used purely depth-based scene understanding in our experiments. This also allows us to study the effect of geometry in isolation.

Beyond semantic understanding, we anticipate that basis scenes will be useful as a standalone dataset for tasks such as benchmarking the performance within Simultaneous Localisation and Mapping in robotics (SLAM) \cite{Handa:etal:ICRA2014} as well as providing environments for reinforcement learning and physical scene understanding with a physics engine \cite{Jiajun:etal:NIPS2015}. All the models in SceneNet are publicly hosted at \url{robotvault.bitbucket.org}.

\subsection{Automatic furniture arrangement}
Past work \cite{Wong:etal:EG15} has focussed on proposing tools to facilitate the annotation of 
meshes or 3D point clouds by human labellers. In addition, free software like Blender\footnote{\url{http://www.blender.org}} 
or CloudCompare\footnote{\url{http://www.danielgm.net/cc/}} can be used to 
manually annotate the objects present in the scene. However, this is a tedious and time-consuming task. This certainly is one limitation with the number of scenes present in SN-BS. Therefore, to add variety in the shapes and categories of objects, we augment SN-BS by generating new physically realistic scenes from object models  downloaded from various online object repositories\cite{Wu:etal:CVPR2015,shapeNet} (see Table \ref{table:repo_sizes}) using simulated annealing \cite{Craigyu:etal:ToG2011}. Most importantly, scene generation is fully automated and sampling objects means that these scenes already come with annotations. 

Different furniture arrangements are optimized with specific feasibility 
constraints \textit{e.g.} support relationships, bounding box constraint, pairwise distances and visibility. We generate scenes in a hierarchical way, imposing constraints first at the object level and then at ``groups of objects" level. Such hierarchical approach is crucial for the optimisation to converge to meaningful configurations when generating cluttered scenes with large number of objects. A similar approach has been proposed in \cite{Merrell:etal:ToG2011} but the system requires user interaction not in keeping with our automatic approach. 

\begin{table}[h]
\centering
\begin{tabular}{|l|c|c|r|}
\hline
\textbf{Repository}  & \textbf{Objects}\\ 
\hline
ModelNet \cite{Wu:etal:CVPR2015}  & 127,915                  \\ \hline
Archive3D & 45,000\\ \hline
Stanford Database \cite{Fisher:etal:SIGGRAPHASIA2012}  & 1,723 \\ \hline
\end{tabular}
\vspace{1mm}
\caption{\small Potential 3D object repositories for scene generation.}
\label{table:repo_sizes}
\end{table}

Inspired by the work of \cite{Merrell:etal:ToG2011} and 
\cite{Craigyu:etal:ToG2011}, we 
formulate \textit{automatic scene generation} from individual objects as an energy 
optimisation problem where the weighted sum of different constraints is 
minimized via simulated annealing. Various constraints and the notations for the 
associated weights and functions are summarised in Table 
\ref{table:energy_constraints}.

\begin{table}[h]
\centering
\begin{tabular}{|l|c|c|c|}
\hline
\textbf{Constraint}  & \textbf{Weight} & \textbf{Function} \\ \hline
Bbox intersection & $w_{bb}$ &  $\max(0,bb_{o,n} - d_{o,n})$ \\ \hline
Pairwise distance & $w_{pw}$ &  $\rho(bb_{o,n},d_{o,n},M_{o,n},\alpha)$ \\ 
\hline
Visibility & $w_{o,n,m}$ & $\nu(v_{o},v_{n},v_{m})$ \\ \hline
Distance to wall  & $w_{o,w}$ &  $\psi(d_{o,w} - d^{\prime}_{o,w})$ \\ \hline
Angle to wall  & $w_{\theta,w}$ &  $\psi(\theta_{o,w} - \theta^{\prime}_{o,w})$ 
\\ \hline
\end{tabular}
\vspace{1mm}
\caption{\small Constraints and notations used for the associated weights and 
functions (see text for details).}
 \label{table:energy_constraints}
\end{table}

In the following, different objects are either denoted by $o$, $m$ or $n$ and the set of all objects is denoted by $\mathcal{O}$. Locations of objects are denoted by $p_o$ such that euclidean distance between two objects is $d_{o,n} = ||p_{o}-p_{n}||_{2}$. The orientations of objects are denoted by $\theta_{o}$. Below, we describe various constraints used in the optimisation

\textbf{Bounding box intersection} A valid configuration of objects must obey 
the very basic criterion of feasibility observed in the real world scenes, 
\textit{i.e.} object bounding boxes must not intersect with each other. We 
denote the bounding box distance $bb_{o,n}$ to be the sum of half 
diagonals of the bounding boxes of the respective objects $o$ and $n$. The distance between 
two objects for any given placement $d_{o,n}$ is the euclidean distance between 
the centres of their bounding boxes. Naturally, $d_{o,n}$ must be greater than 
or equal to $bb_{o,n}$ for a placement to be feasible. Any deviation 
from this constraint is penalized by $\max(0,bb_{o,n}-d_{o,n})$ \cite{Craigyu:etal:ToG2011}.     

\begin{figure}[t!]
\centerline{
\includegraphics[width=0.8\linewidth]{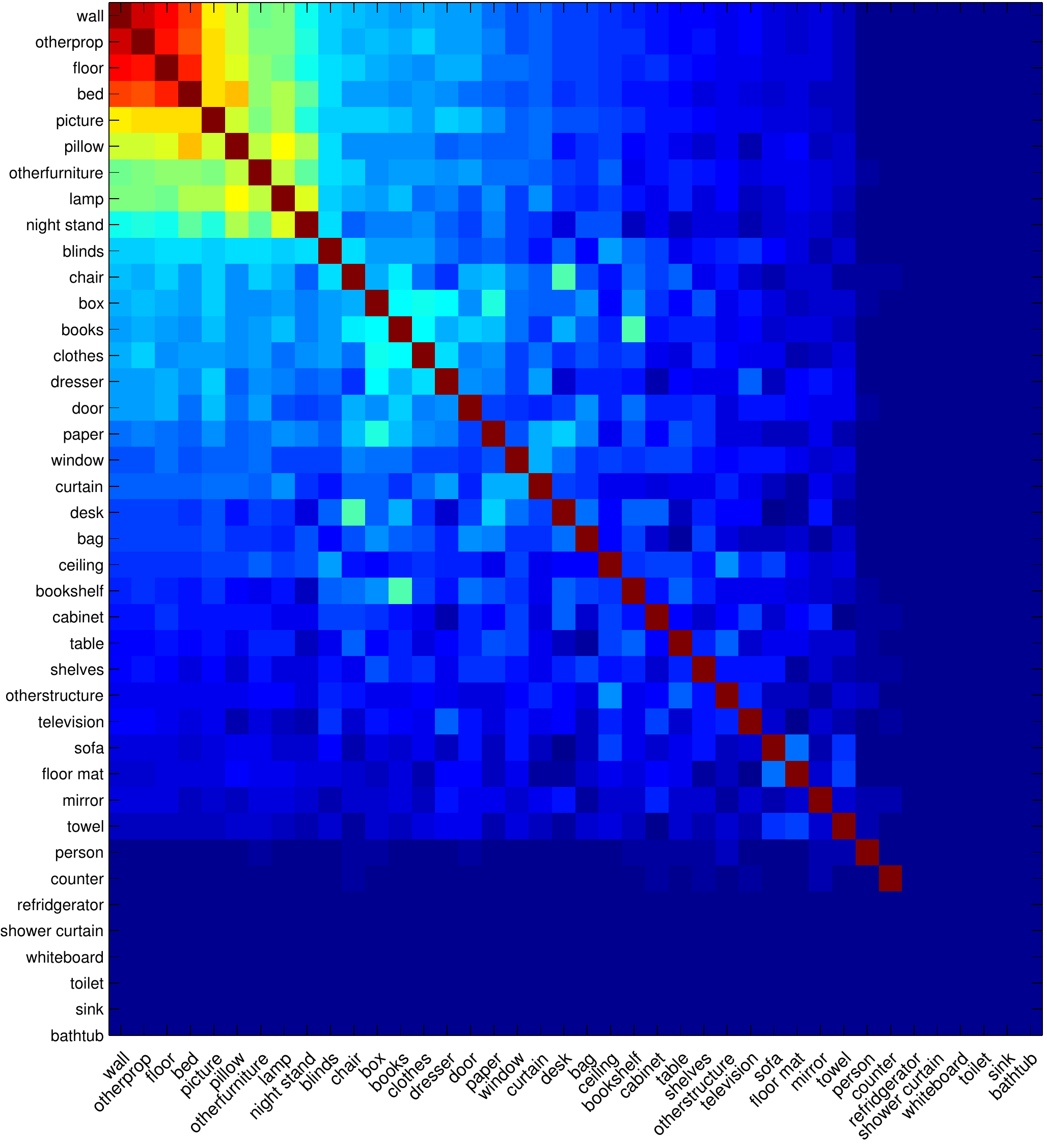}
}
\caption{\small
Co-occurrence statistics for bedroom scenes in NYUv2 40 
class labels. Warmer colours reflect higher co-occurrence frequency.} 
%\hrule
\label{fig:nyu_beds}
\end{figure}

\textbf{Pairwise distances} Using the statistics extracted from man-made scenes 
in NYUv2 (see Fig.\ref{fig:nyu_beds}) and SN-BS, objects that are 
more likely to co-occur are paired together, \textit{e.g.} nightstands are 
likely to appear next to beds, chairs next to tables, monitors on the desk 
\textit{etc.} capturing the contextual relationships 
between objects. We use a slight variation of the pairwise term used in 
\cite{Merrell:etal:ToG2011}
\[ \rho(bb_{o,n},d_{o,n},M_{o,n},\alpha) =
  \begin{cases}
    (\frac{bb_{o,n}}{d_{o,n}})^{\alpha} & \text{if } d_{o,n} < bb_{o,n} \\
    0  & \text{if } bb_{o,n} < d_{o,n} < M_{o,n}\\
    (\frac{d_{o,n}}{M_{o,n}})^{\alpha} & \text{if } d_{o,n} > M_{o,n}
  \end{cases}
\]
where $M$ is the maximum recommended distance between $o$ and $n$ \cite{Merrell:etal:ToG2011}. In our 
experiments we have found that $\alpha=2$ works reasonably well. Different 
pairwise constraints that frequently appear in our experiments are between \textit{beds and cupboards, beds and nightstands, chairs and tables, tables and tv, and desks and chairs}.

\textbf{Visibility constraint} Visibility constraint ensures that one 
object is fully visible from the other along the ray joining their centres \textit{e.g.} TV should be visible from sofa with no other object in the view and therefore, bounding box intersection of any other object is penalised with the combined bounding box of TV and sofa. Our visibility term is inspired from \cite{Craigyu:etal:ToG2011} and is defined as 
\begin{eqnarray}
\nu(v_{o},v_{n},v_{m}) = \sum_{m=1}^{N} w_{on,m} \max(0,bb_{on,m}-d_{on,m})
\end{eqnarray}
where $bb_{on,m}$ is the sum of the half diagonal of the bounding box of $m$ 
and the diagonal of the combined bounding box of $o$ and $n$, while 
$d_{on,m}$ is the euclidean distance between the corresponding bounding boxes.

\textbf{Distance and angle with wall} Many objects in the indoor scenes are more likely to occur rested against walls \textit{e.g.} beds, cupboards and desks. We add another prior term to increase the likelihood of such objects satisfying this behaviour. For each object, we denote euclidean distance between the centre of the bounding box of the object and the nearest wall by $d_{o,w}$ and the corresponding prior by $d^{\prime}_{o,w}$\cite{Craigyu:etal:ToG2011}. Similarly for angle we have $\theta_{o,w}$ and $\theta^{\prime}_{o,w}$ respectively. Our distance and angle penalties are standard $\mathcal{L}_{2}^{2}$ terms $\psi(x) = x^{2}$.

\begin{figure}[t]
\centerline{
\includegraphics[width=0.285\linewidth]{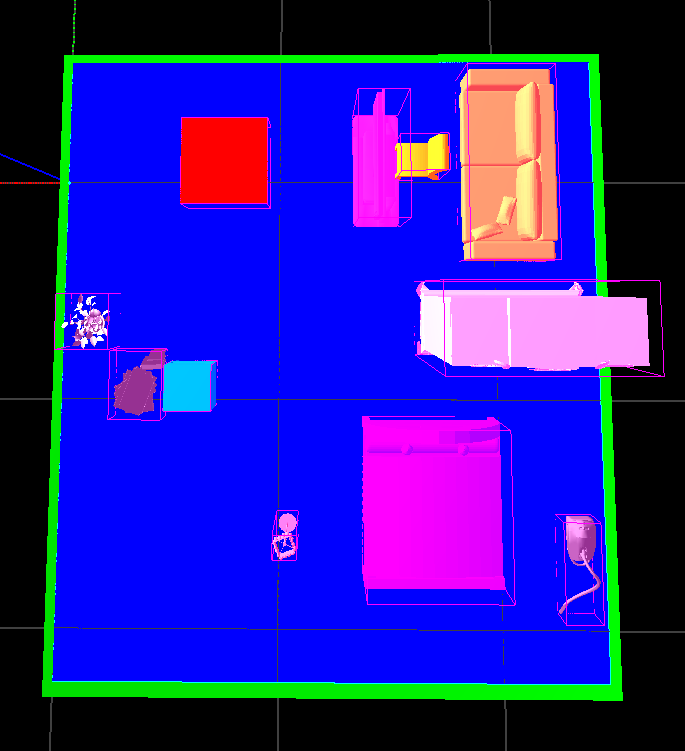}
\includegraphics[width=0.310\linewidth]{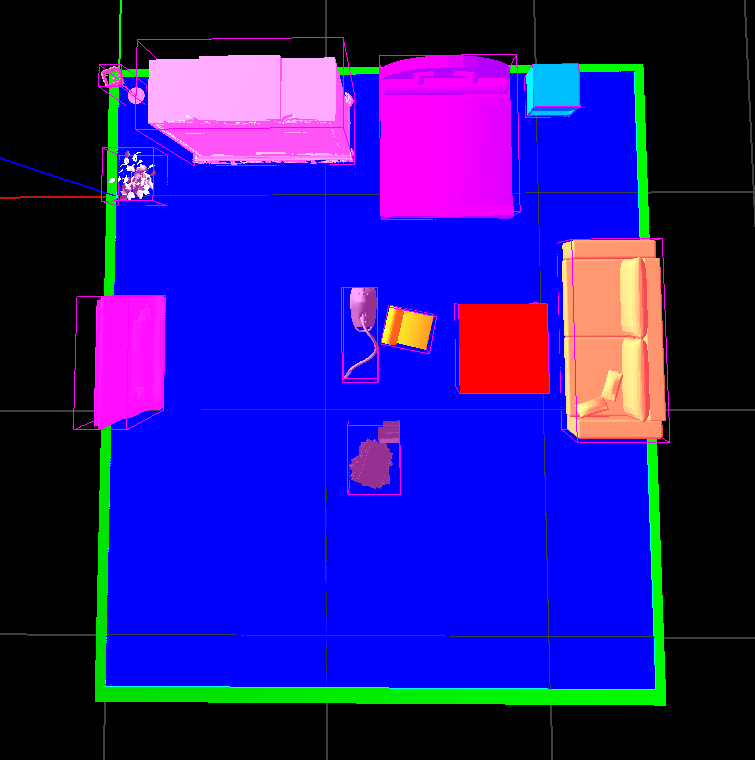}
\includegraphics[width=0.3\linewidth]{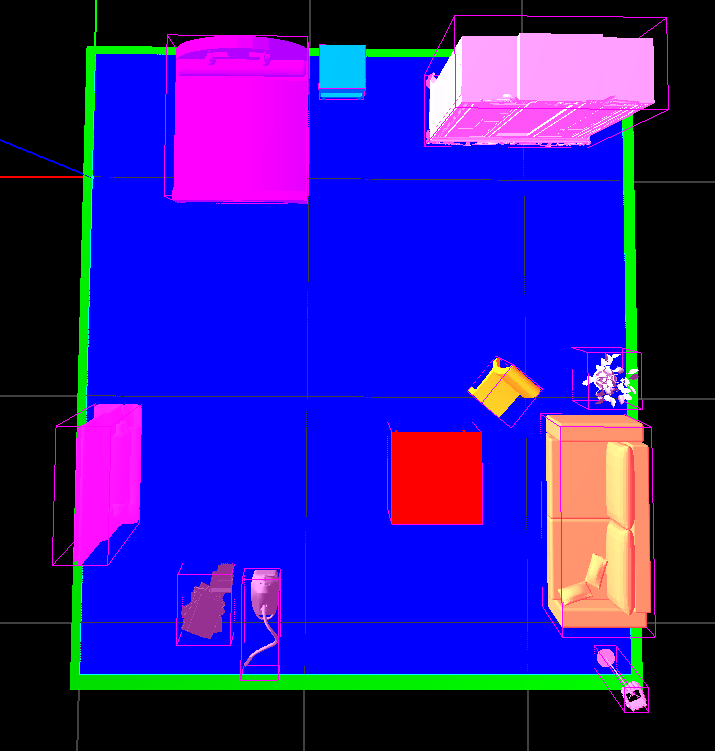}
}
\centerline{
\hfill
\makebox[0.3\linewidth][c] { \footnotesize{(a) No pairwise/visibility}}
\hfill
\makebox[0.3\linewidth][c] { \footnotesize{(b) No visibility} }
\hfill
\makebox[0.33\linewidth][c] { \footnotesize{(c) All constraints}}
}
\caption{\small
Effect of different constraints on the optimisation. With no pairwise or visibility constraints, objects appear scattered at random (a). When pairwise constraints are added, the sofa, table and TV assume sensible relative positions but with chair and vacuum cleaner occluding the view (b). With all constraints, occlusions are removed.} \vspace{0.5mm} 
\label{fig:sim_scenes_constraints}
\end{figure}

The overall energy function is then the weighted sum of all the constraints and minimised via simulated annealing to optimise over the positions and angles of each object $\mathcal{C}=\{p_{o},\theta_{o}\}$. An illustration of how different constraints affect the optimisation is given in Fig.\ref{fig:sim_scenes_constraints}.
\begin{align}
\mathcal{E}(\mathcal{C}) = \sum_{o \in \mathcal{O} } \bigg\{ \sum_{n \in \mathcal{O} } & 
\big\{ w_{bb} \max(0,bb_{o,n} - d_{o,n}) \notag\\ 
& + w_{pw} \rho(bb_{o,n},d_{o,n},M_{o,n},\alpha) \notag\\
& + w_{\theta} \psi(\theta_{o,n} - \theta^{\prime}_{o,n}) \notag\\
& + \sum_{m \in \mathcal{O}} w_{o,n,m} \max(0,bb_{on,m}-d_{on,m})\big\} \notag\\
& + w_{o,w} \psi(d_{o,w} - d^{\prime}_{o,w}) \notag\\
& + w_{\theta} \psi(\theta_{o,w} - \theta^{\prime}_{o,w}) \bigg\}
\end{align}
Note that we also used angle prior between different objects and the corresponding terms are denoted by $\theta_{o,n}$ and $\theta^{\prime}_{o,n}$ respectively. 
\subsection{Adding noise to ground truth data}
\label{section:addnoise}
\begin{figure*}
\centerline{
\includegraphics[width=0.33\linewidth]{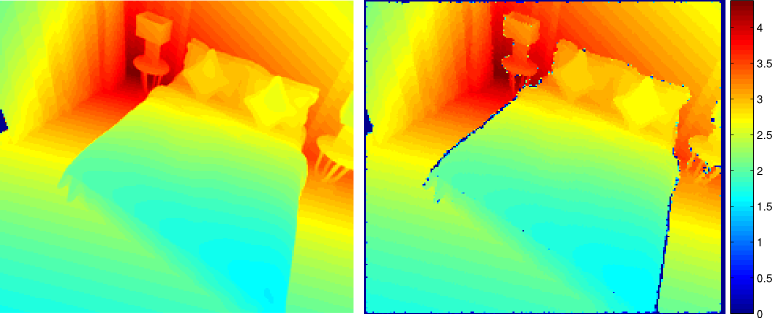}
\includegraphics[width=0.33\linewidth]{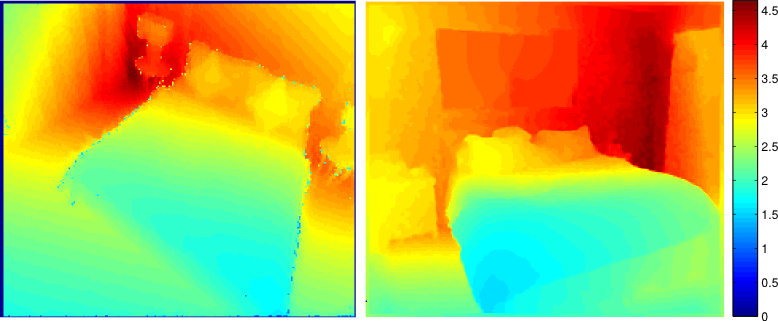}
\includegraphics[width=0.33\linewidth]{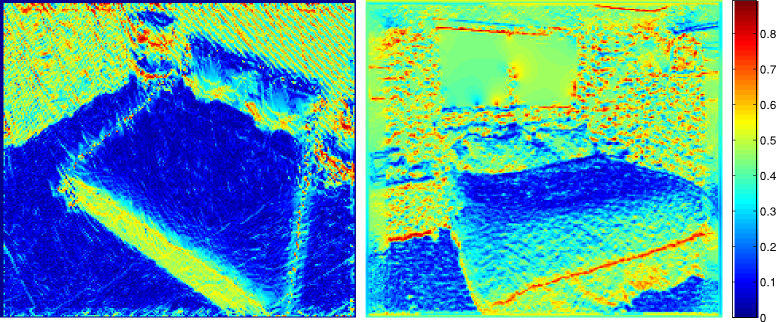}
}
\centerline{
\hfill
\makebox[0.3\linewidth][c] { \footnotesize{(a) noisy depth}}
\hfill
\makebox[0.3\linewidth][c] { \footnotesize{(b) depth denoised } }
\hfill
\makebox[0.33\linewidth][c] { \footnotesize{(c) angle with gravity}}
}
\caption{\small
Left image in (a) shows the perfect rendered depth map and right image shows the corresponding noisy image provided by our simulator. In (b) we show a side by side comparison of the inpainted depth maps of one of the bedroom scenes in SceneNet with a similar looking bedroom in NYUv2 test data and (c) shows the comparison of angle with gravity vector images. Importantly, left image in (c) highlights the view point invariance of the angle with gravity.} 
\label{fig:SceneNet-NYU comparison}
\end{figure*}

We realise the possible mismatch of the distribution of the noise characteristics in real world datasets and our synthetic depth-maps and therefore add noise to the perfect rendered depth-maps according to the simulated kinect noise model in \cite{Gschwandtner:etal:AVC2011,Handa:etal:ICRA2014} (see Fig.  \ref{fig:SceneNet-NYU comparison} for visualisations). This is to ensure that any model trained on synthetic datasets can have a significant effect on real world depth data either directly or via fine-tuning \cite{Fischer:etal:arXiv2015}.

%\section{Experimental Set-up}
Our primary goal in this work is to show how synthetic data can enable improvements in per-pixel semantic segmentation on real world indoor scenes. For all our experiments, we choose the state-of-the-art segmentation algorithm \cite{Noh:etal:ARXIV2015,Badrinarayanan:etal:ARXIV2015} with encoder-decoder architecture built on top of the popular VGG network \cite{Simonyan:etal:ARXIV2014} and use the publicly available code of \cite{Noh:etal:ARXIV2015}. Both algorithms have only been used for RGB image based segmentation, therefore, we adapted them to work on depth based three channel input, DHA,\footnote{A similar HHA encoding has been used in \cite{Gupta:etal:CVPR2015} but like \cite{Eigen:etal:ICCV2015} we did not observe much difference.} namely depth, height from ground plane and angle with gravity vector. Since we already know the camera poses when rendering, it is straightforward to obtain height and angle with gravity vector for synthetic data but we implemented a C++ GPU version of the otherwise slow MATLAB code of \cite{Gupta:etal:CVPR2013} to align the NYUv2 depth-maps to intertial frame to obtain the corresponding heights and angles. SUN RGB-D already provide the tilt angle and rotation matrices so we use them to obtain our DHA features. We intially report results on only 11 different categories (see Table \ref{table: CA breakdown}). This is because, to generate new scenes, we sample objects from axis aligned ModelNet10 \cite{Wu:etal:CVPR2015} which does not contain \textit{painting} and \textit{books} that in total add to the 13 classes used in \cite{Eigen:etal:ICCV2015,Long:etal:CVPR2015}. However, we take special care in doing comparisons with \cite{Eigen:etal:ICCV2015} who made their per-pixel semantic predictions for the NYUv2 test data publicly available. We later, also report results on 13 classes by directly finetuning on the standard 13 class datasets.

Also, we inpainted the noisy depth-maps from our simulator with the MATLAB colorization code provided in the NYUv2 dataset. This was to ensure that our depth-maps qualitatively match with the inpainted depth-maps in NYUv2 (see Fig.\ref{fig:SceneNet-NYU comparison}). We tried the colorization with different kernel sizes and empirically found 3$\times$3 to be good enough --- our final results did not change much with 7$\times$7 and bigger kernels.

\section{Results}
We collect a wide variety of random images sampled from SceneNet and convert each depth image into three channel DHA input. Our network is first trained on purely synthetic data and then fine tuned on the 795 training images provided in the NYUv2 dataset. Finally, NYUv2 test data is used to compare the results with different variants of training data used in experiments.

We denote training performed on noise injected synthetic data from our repository, SceneNet, as SceneNet-DHA, and fine tuning on training data in NYUv2 by SceneNet-FT-NYU-DHA. When using dropout we denote them by SceneNet-DO-DHA and SceneNet-FT-NYU-DO-DHA respectively. Similarly, the networks trained on NYU are denoted by NYU-DHA and NYU-DO-DHA. We also tried dropout at test time as mentioned in \cite{Gal:etal:arXiv2015} but this comes at the expense of increased inference time --- dropout at test time is shown to have similar effect of multiple model averaging and at the same time provides uncertainty in the final predictions. First, we quantify the performance of all these variants on the NYUv2 test data and later on a bigger real world dataset, SUN RGB-D, based on standard global and class accuracy metrics. 

We use DHA images of size 224$\times$224 as input to the network and intialise the weights with pretrained VGG network. At first, it may seem that the network trained on images cannot be useful for training on depth data, however, since the first layer filters often look for edges and orientations \cite{Zeiler:etal:ECCV2014}, it is sensible to use them for relatively similar modalities like depth --- edges in depth almost always align with the RGB edges --- and deeper layers adapt accordingly. Therefore, in all our experiments the network always converged with accuracies in high nineties on the training set. Also, it is worth noting that FCN \cite{Long:etal:CVPR2015} quantise the HHA to $[$0,255$]$ for an equivalent RGB image and use pretrained VGG network to initialise. We do not quantise and maintain the floating point precision of the DHA images.
\begin{figure*}[htp]
\centerline{
\includegraphics[width=0.20\linewidth]{nyu_img_rgb_0504.png}
\includegraphics[width=0.20\linewidth]{nyu_img_rgb_0207.png}
\includegraphics[width=0.20\linewidth]{nyu_img_rgb_0523.png}
\includegraphics[width=0.20\linewidth]{nyu_img_rgb_0651.png}
\includegraphics[width=0.20\linewidth]{nyu_img_rgb_0611.png}
}
\centerline{
\includegraphics[width=0.20\linewidth]{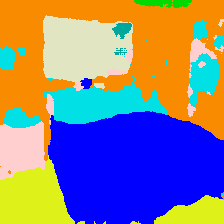}
\includegraphics[width=0.20\linewidth]{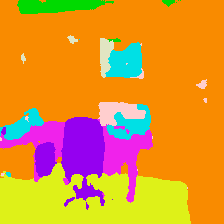}
\includegraphics[width=0.20\linewidth]{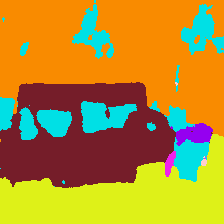}
\includegraphics[width=0.20\linewidth]{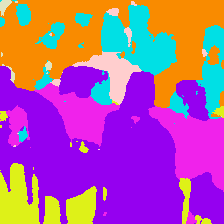}
\includegraphics[width=0.20\linewidth]{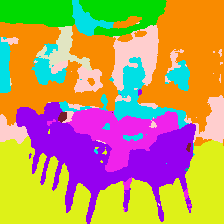}
}
\centerline{
\includegraphics[width=0.20\linewidth]{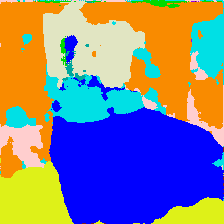}
\includegraphics[width=0.20\linewidth]{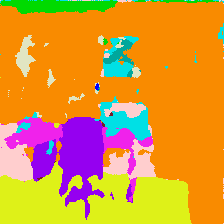}
\includegraphics[width=0.20\linewidth]{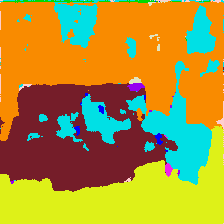}
\includegraphics[width=0.20\linewidth]{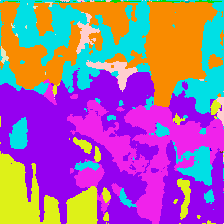}
\includegraphics[width=0.20\linewidth]{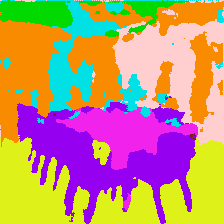}
}
\centerline{
\includegraphics[width=0.20\linewidth]{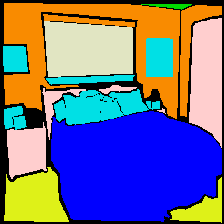}
\includegraphics[width=0.20\linewidth]{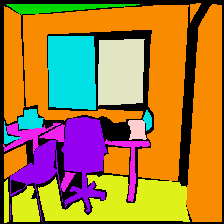}
\includegraphics[width=0.20\linewidth]{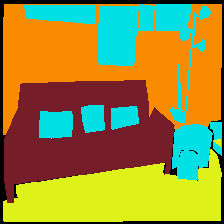}
\includegraphics[width=0.20\linewidth]{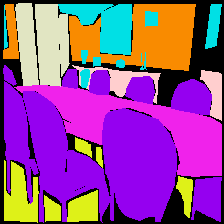}
\includegraphics[width=0.20\linewidth]{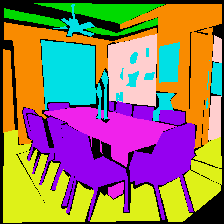}
}
\caption{\small
Results for 11 classes on NYUv2 test data obtained with SceneNet-FT-NYU-DHA and NYU-DHA. First row shows the RGB images, predictions returned by SceneNet-FT-NYU-DHA and NYU-DHA are shown in second and third row respectively. Last row shows the ground truth.} \vspace{1mm}
\label{fig:NYUv2results}
\end{figure*}

\begin{figure}
\centerline{
\includegraphics[width=\linewidth]{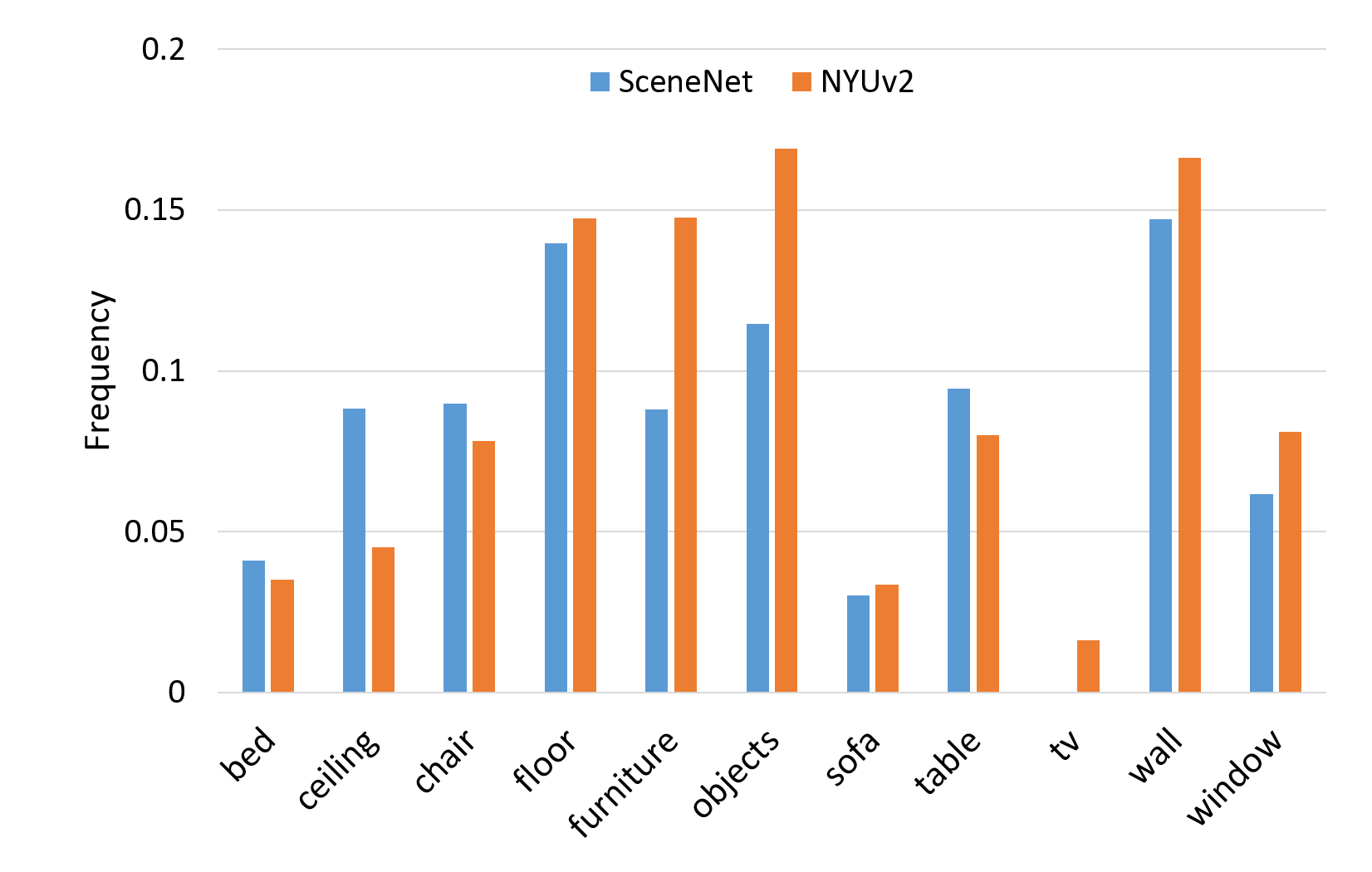}
}
\caption{\small
Side by side comparison of proportions of objects in the 9K images rendered with SceneNet and 795 training images in NYUv2.} 
\label{fig:freq_SceneNet_NYU}
\end{figure}

We render 10,030 depth images from random view points ensuring that a minimum number of objects is visible --- avoiding camera looking at only walls or floor --- and perturb the depth values according to Section \ref{section:addnoise} to generate depth maps qualitatively similar to NYUv2. Comparison of different proportions of objects in these rendered images and NYUv2 training data is shown Fig. \ref{fig:freq_SceneNet_NYU}. All the models are trained with stochastic gradient descent with a starting learning rate of 0.01 which is multiplied by 0.9 every 3-4 epochs, a weight decay of 0.0005, and a momentum of 0.9. We characterise the experiments into comparisons with real training data in NYUv2 and the state of the art, Eigen and Fergus \cite{Eigen:etal:ICCV2015}, for both 11 and 13 class segmentations. We also perform experiments on SUNRGBD for 13 classes and set a new benchmark for pure depth based segmentation.

\begin{table}[h]
\centering
\begin{tabular}{l}
\textbf{11 Class Semantic Segmentation: NYUv2} \\ 
\end{tabular}
\begin{tabular}{|l|p{1.2cm}|p{1.2cm}|}
\hline
Training & global acc. & class acc. \\ \hline
%Hermans \textit{et al.}(rgbd+crf) \cite{Hermans:etal:ICRA2014} & -- & 49.4 \\ \hline
SceneNet-DHA & 54.4 & 42.6 \\ \hline
NYU-DHA & 63.8 & 53.7 \\ \hline
SceneNet-FT-NYU-DHA & 67.4 & 59.1 \\ \hline
SceneNet-DO-DHA & 54.6 & 43.2 \\ \hline
NYU-DO-DHA & 65.7 & 56.9 \\ \hline
SceneNet-FT-NYU-DO-DHA & 68.0 & 59.9 \\ \hline
Eigen \textit{et al.}(rgbd+normals) \cite{Eigen:etal:ICCV2015} & \textbf{69.5} & \textbf{63.0} \\ \hline
\end{tabular}\vspace{0.7mm}
\caption{Different variants of training data that we use in our experiments. The performance jump from NYU-DHA to SceneNet-FT-NYU-DHA is clear. Adding dropout helps most in the NYU-DO-DHA but SceneNet-FT-NYU-DO-DHA shows only a marginal improvement. Overall increase in performance from NYU-DO-DHA to SceneNet-FT-NYU-DO-DHA is by 2.3\% and 3\% in global and class accuracy respectively. Note that we recomputed the accuracies of \cite{Eigen:etal:ICCV2015} using their publicly available annotations of 320$\times$240 and resizing them to 224$\times$224. Note that Hermans \textit{et al.} \cite{Hermans:etal:ICRA2014} predictions are not publicly available. As a result, we cannot evaluate their performance on these 11 classes.} 
\label{table: comparison of various networks on GA and CA}
\end{table}

\paragraph{Comparisons with NYUv2 training} Training purely on synthetic data, SceneNet-DHA, results in only modest performance on the test data (see Table \ref{table: comparison of various networks on GA and CA}). Comparison with NYU-DHA reveals that fine tuning is needed to obtain further improvements in the results. As a result, we see that the performance jump from NYU-DHA to SceneNet-FT-NYU-DHA is very clear --- an increase of 5.4\% in the class and 3.6\% in the global accuracy showing the usefulness of synthetic data for real world scene segmentation. Importantly, convergence was twice as fast for SceneNet-FT-NYU-DHA compared to NYU-DHA. Specifically, prior to fine-tuning we trained SceneNet-DHA for 21K iterations (14 epochs). Fine tuning on NYUv2 took another 4K iterations (30 epochs) while NYU-DHA took about 7.5K iterations (56 epochs) to converge. Qualitative results of the segmentation are shown in Fig. \ref{fig:NYUv2results} which highlights the impact of synthetic data in the final results.
\paragraph{Comparisons with Eigen and Fergus \cite{Eigen:etal:ICCV2015}} We also compare our results to the state-of-the-art systems of Eigen \textit{et al.}\cite{Eigen:etal:ICCV2015} who use data augmentation and a combination of RGBD and normals. Since we use only depth, our system is not directly comparable to \cite{Eigen:etal:ICCV2015} but we obtain competitive performance. SceneNet-FT-NYU-DHA although performs better than NYU-DHA, it still falls short of the performance by state of the art method \cite{Eigen:etal:ICCV2015}. However, careful examination in class accuracies (see Table \ref{table: CA breakdown}) reveals that we perform comparably only compromising on \textit{tv} and \textit{windows} --- we expect RGB to play a bigger role here --- emphasizing that for functional categories of objects geometry is a strong cue for segmentation. We used the publicly available annotations of \cite{Eigen:etal:ICCV2015} to re-evaluate the accuracies for the 11 classes we use.
\begin{table*}[t]
\centering
\begin{tabular}{l}
\textbf{11 Class Semantic Segmentation: NYUv2} \\ 
\end{tabular}
%\begin{tabular}{|l|p{0.5cm}|p{0.5cm}|p{0.5cm}|p{0.5cm}|p{0.5cm}|p{0.5cm}|p{0.5cm}|p{0.5cm}|p{0.5cm}|p{0.5cm}|p{0.5cm}|p{0.5cm}|}
\begin{tabular}{|l|l|l|l|l|l|l|l|l|l|l|l|l|}
\hline
Training  & \cellcolor{bedColor}{bed}    & \cellcolor{ceilColor}{ceil.} & \cellcolor{chairColor}{chair}  & \cellcolor{floorColor}{floor}  & \cellcolor{furnColor}{furn}   & \cellcolor{objsColor}{objs.} & \cellcolor{sofaColor}{sofa}   & \cellcolor{tableColor}{table}  & \cellcolor{tvColor}{tv}     & \cellcolor{wallColor}{wall}   & \cellcolor{windColor}{window} \\ \hline
NYU-DHA & 64.5 & 68.2 & 51.0 & 95.0 & 51.0 & 48.2          & 49.7 & 41.9 & 12.8 & 84.2 & 24.5 \\ \hline
SceneNet-DHA & 60.8 & 43.4 & 68.5 & 90.0 & 26.5 & 24.3 &  21.2 & 42.1 & 0 & \textbf{92.1} & 0.3 \\ \hline
SceneNet-FT-NYU-DHA & 70.3  & 75.9 & 59.8 & 96.0  & \textbf{60.7} & 49.7 &  59.9 & \textbf{49.7} & 24.3 &  84.8  & 27.9 \\ \hline
NYU-DO-DHA & 69.0 & 74.6 & 54.0 & 95.6 & 57.1 & 48.7 &55.7 & 42.5 & 18.5 & 84.7 & 25.5 \\ \hline
SceneNet-DO-DHA & 67.7 & 40.9 & 67.5 & 87.8 & 38.6 & 22.6 & 15.8 & 44.2 & 0 & 89.0 & 0.8 \\ \hline
SceneNet-FT-NYU-DO-DHA & \textbf{72.5} & 74.1 & 61.0 & 96.2 & 60.4 & 50.0 & \textbf{62.8} & 43.8 & 19.4 & 85.3 & 30.0 \\ \hline
Eigen \textit{et al.} (rgbd+normals) \cite{Eigen:etal:ICCV2015} & 61.1 & 78.3 & \textbf{72.1} & \textbf{96.5} & 
55.1 & \textbf{52.1} & 45.8 & 45.0 & \textbf{41.9} & 88.7 & \textbf{57.7} \\ \hline
%Hermans \textit{et al.}(rgbd+crf)\cite{Hermans:etal:ICRA2014} & 68.4 & \textbf{83.4} & 41.9 & 91.5 & 37.1 & 8.6 & 28.5 & 27.7 & 38.4 & 71.8 & 46.1 \\ \hline
%SceneNet-FT-NYU-DHA(1.5k) & 54.19& 62.3 & 38.1 & 73.3 & 47.0 & 37.5 & 43.6 & 27.5 & 16.7 & 71.1 & 17.4 \\ \hline
\end{tabular}
\vspace{0.5mm} \vspace{0.5mm}
\caption{Different training variants on the NYUv2 test set. The benefits of adding synthetic data are clear. For fair comparison, it should ne noted that \cite{Eigen:etal:ICCV2015} use RGBD+Normals vs depth alone and \cite{Hermans:etal:ICRA2014} use RGBD and a CRF to smooth the results. We would like to stress here that our network is trained end-to-end as compared to multi-stage training done in Eigen \textit{et al.} \cite{Eigen:etal:ICCV2015} and does not use any RGB data or data augmentation. We outperform \cite{Eigen:etal:ICCV2015} on \textit{sofas} and \textit{beds} but fall behind on \textit{chairs}, \textit{tv} and \textit{windows}. We expect \textit{tv} and \textit{windows} are likely to be segmented better with RGB data. However, we obtain comparable performance on rest of the  classes further emphasising that for functional categories of objects shape is a strong cue. Poor performance of SceneNet-DHA and SceneNet-DO-DHA on \textit{tv} and \textit{windows} is mainly due to limited training data for these classes in SceneNet.}
\label{table: CA breakdown}
\end{table*}
\paragraph{Comparisons with dropout} We also used dropout ratio of 0.5 in the fully connected layers to gauge the effect of regularisation in the parameters to prevent over-fitting. We see 3.2\% improvement in class and 1.9\% global accuracy in the results with real training data used in NYU-DO-DHA compared to NYU-DHA. However, we only observed minor improvements with both SceneNet-DO-DHA and SceneNet-FT-NYU-DO-DHA against SceneNet-DHA and SceneNet-FT-NYU-DHA respectively suggesting that increase in data acts as an implicit regulariser. Further improvements depend largely on the amount of training data used \cite{AdamCoatesLecture}. Although there is no limit to the amount of training data we can render we are only limited by GPU speed and memory --- training time for a batch size of 6 images takes about 0.8s for forward pass and 1.2s for backward pass, a combined total of 2s per iteration on NVidia Tesla K80. We also anticipate that training can be made efficient by either completely forgoing the fully connected layers \cite{Badrinarayanan:etal:ARXIV2015} or reducing their number of features \cite{Chen:etal:ICLR2014} --- fully connected layers in VGG contain nearly $\frac{204800}{246912}\backsim 83\%$ of the parameters. Training on data of the order of ImageNet \cite{Deng:etal:CVPR2009} remains an exciting opportunity for the future. 

We also used dropout at test time \cite{Gal:etal:arXiv2015} but observed very similar performance gain without it. However, dropout at test time \cite{Gal:etal:arXiv2015} makes the network robust to out-of-domain data. We leave it as an interesting future direction to explore. Overall, we have SceneNet-FT-NYU-DO-DHA as a clear winner against NYU-DO-DHA as shown in Table \ref{table: comparison of various networks on GA and CA}. 

\begin{figure*}[htp]
\centerline{
\includegraphics[width=0.20\linewidth]{nyu_img_rgb_0590.png}
\includegraphics[width=0.20\linewidth]{nyu_img_rgb_0624.png}
\includegraphics[width=0.20\linewidth]{nyu_img_rgb_0442.png}
\includegraphics[width=0.20\linewidth]{nyu_img_rgb_0413.png}
\includegraphics[width=0.20\linewidth]{nyu_img_rgb_0237.png}
}
\centerline{
\includegraphics[width=0.20\linewidth]{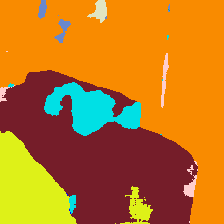}
\includegraphics[width=0.20\linewidth]{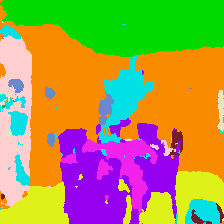}
\includegraphics[width=0.20\linewidth]{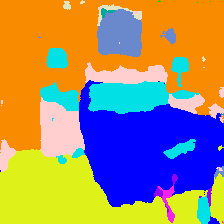}
\includegraphics[width=0.20\linewidth]{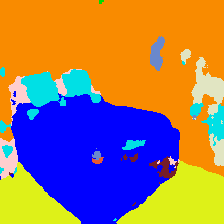}
\includegraphics[width=0.20\linewidth]{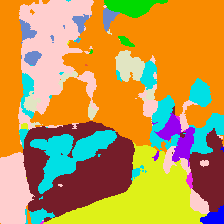}
}
\centerline{
\includegraphics[width=0.20\linewidth]{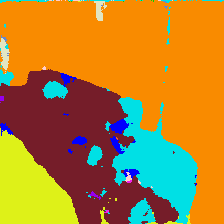}
\includegraphics[width=0.20\linewidth]{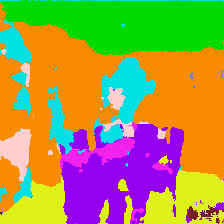}
\includegraphics[width=0.20\linewidth]{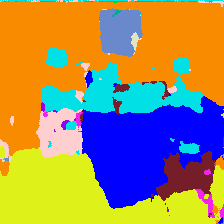}
\includegraphics[width=0.20\linewidth]{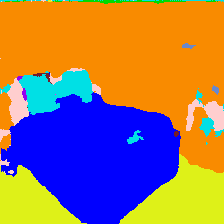}
\includegraphics[width=0.20\linewidth]{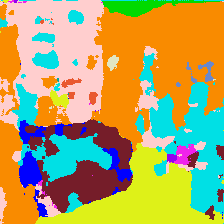}
}
\centerline{
\includegraphics[width=0.20\linewidth]{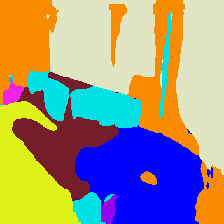}
\includegraphics[width=0.20\linewidth]{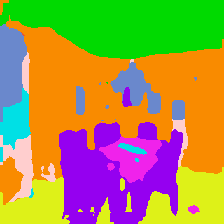}
\includegraphics[width=0.20\linewidth]{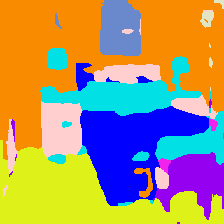}
\includegraphics[width=0.20\linewidth]{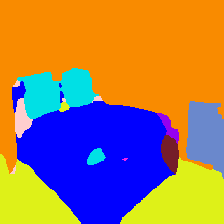}
\includegraphics[width=0.20\linewidth]{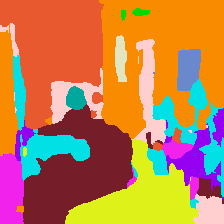}
}
\centerline{
\includegraphics[width=0.20\linewidth]{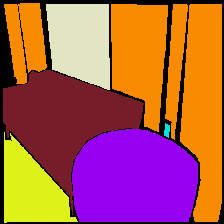}
\includegraphics[width=0.20\linewidth]{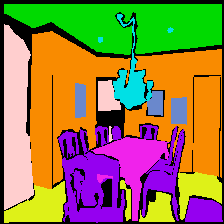}
\includegraphics[width=0.20\linewidth]{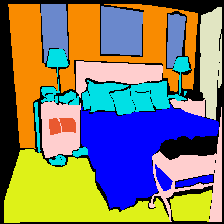}
\includegraphics[width=0.20\linewidth]{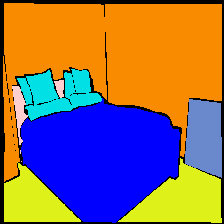}
\includegraphics[width=0.20\linewidth]{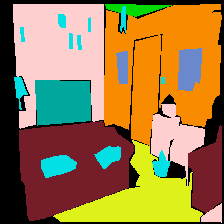}
}
%\centerline{
%\hfill
%\makebox[0.20\linewidth][c] { \footnotesize{(a) 504}}
%\hfill
%\makebox[0.20\linewidth][c] { \footnotesize{(b) 207}}
%\hfill
%\makebox[0.20\linewidth][c] { \footnotesize{(c) 523}}
%\hfill
%\makebox[0.20\linewidth][c] { \footnotesize{(d) 651}}
%\hfill
%\makebox[0.20\linewidth][c] { \footnotesize{(e) 611}}
%}
\caption{\small
Results for 13 class labels: First row is the RGB image. Second row shows the results obtained with SceneNet-FT-NYU-DHA while the third row shows the results with NYU-DHA. Fourth row has the predictions returned by Eigen and Fergus \cite{Eigen:etal:ICCV2015}. Last row shows the ground truth. Looking at the results, it is evident that the predictions by \cite{Eigen:etal:ICCV2015} tend to be smoother and contained unlike NYU-DHA as well as SceneNet-FT-NYU-DHA. We believe this difference is a result of rich input channels in RGB, normals and depth used by \cite{Eigen:etal:ICCV2015} while we use only depth in our experiments.} \vspace{1mm}
\label{fig:NYUv2results: labels13}
\end{figure*}

\paragraph{Results on 13 Class Semantic Segmentation: NYUv2} We also performed our experiments on the 13 class semantic segmentation task. It is worth remembering that the two extra classes we add to the 11 class experiment are \textit{painting} and \textit{books}. 
Although, we are limited by the number of labels for \textit{painting} and \textit{books} in SceneNet, we fine tune directly on 13 class semantic segmentation task on NYUv2. The performance gain from NYU-DHA to SceneNet-FT-NYU-DHA is evident, highlighting the role of synthetic data. As seen in Table \ref{table: CA breakdown for 13 classes}, SceneNet-FT-NYU-DHA performs consistently better than NYU-DHA for all the classes. We observe similar trend for the comparison between NYU-DO-DHA and SceneNet-FT-NYU-DO-DHA. It is worth remembering that Eigen and Fergus \cite{Eigen:etal:ICCV2015} use RGB and normals together with the depth channel and hence maintain superior overall performance over our methods that use only depth data. Again, careful examination reveals that we only compromise on \textit{books}, \textit{painting}, \textit{tv} and \textit{windows} --- we expect RGB to play a bigger role here in segmenting these classes. This is also reflected in the overall mean global and class accuracies as shown in Table \ref{table: comparison of various networks on GA and CA for 13 classes}. Figures \ref{fig:NYUv2results: labels13} compares predictions returned by SceneNet-FT-NYU-DHA, NYU-DHA and Eigen and Fergus \cite{Eigen:etal:ICCV2015} on a variety of test images in NYUv2 dataset.

\begin{table*}
\begin{tabular}{l}
\textbf{13 Class Semantic Segmentation: NYUv2} \\ 
\end{tabular}
\centering
\begin{tabular}{|l|p{0.5cm}|p{0.5cm}|p{0.5cm}|p{0.5cm}|p{0.5cm}|p{0.5cm}|p{0.5cm}|p{0.5cm}|p{0.5cm}|p{0.5cm}|p{0.5cm}|p{0.5cm}|p{0.5cm}|p{0.5cm}|}
\hline
Training  & \cellcolor{bedColor}\rotatebox{90}{bed} & \cellcolor{booksColor}\rotatebox{90}{books}  & \cellcolor{ceilColor}\rotatebox{90}{ceil.} & \cellcolor{chairColor}\rotatebox{90}{chair}  & \cellcolor{floorColor}\rotatebox{90}{floor}  & \cellcolor{furnColor}\rotatebox{90}{furn}   & \cellcolor{objsColor}\rotatebox{90}{objs.} & \cellcolor{paintColor}\rotatebox{90}{paint.} & \cellcolor{sofaColor}\rotatebox{90}{sofa}   & \cellcolor{tableColor}\rotatebox{90}{table}  & \cellcolor{tvColor}\rotatebox{90}{tv}     & \cellcolor{wallColor}\rotatebox{90}{wall}   & \cellcolor{windColor}\rotatebox{90}{window} \\ \hline
NYU-DHA & 67.7 & 6.5 & 69.9 & 47.9 & \textbf{96.2} & 53.8 & 46.5 & 11.3 & 50.7 & 41.6 & 10.8 & 85.0 & 25.8 \\ \hline
SceneNet-DHA & 60.8 & 2.0 & 44.2 & 68.3 & 90.2 & 26.4 & 27.6 &  6.3 & 21.1 & 42.2 & 0 & \textbf{92.0} & 0.0 \\ \hline
SceneNet-FT-NYU-DHA & 70.8 & 5.3 & 75.0 & 58.9 & 95.9 & 63.3 & 48.4 & 15.2 & 58.0 & 43.6 & 22.3 &  85.1  & 29.9 \\ \hline
NYU-DO-DHA & 69.6 & 3.1 & 69.3 & 53.2 & 95.9 & 60.0 & 49.0 & 11.6 & 52.7 & 40.2 & 17.3 & 85.0 & 27.1 \\ \hline
SceneNet-DO-DHA & 67.9 & 4.7 & 41.2 & 67.7 & 87.9 & 38.4 & 25.6 &  6.3 & 16.3 & 43.8 & 0 & 88.6 & 1.0 \\ \hline
SceneNet-FT-NYU-DO-DHA & \textbf{70.8} & 5.5 & 76.2 & 59.6 & 95.9 & \textbf{62.3} & \textbf{50.0} & 18.0 & \textbf{61.3} & 42.2 & 22.2 & 86.1 & 32.1 \\ \hline

Eigen \textit{et al.} (rgbd+normals) \cite{Eigen:etal:ICCV2015} & 61.1 & \textbf{49.7} & 78.3 & \textbf{72.1} & 96.0 & 55.1 & 40.7 &\textbf{58.7} & 45.8 &\textbf{44.9}& \textbf{41.9} & 88.7 & \textbf{57.7}  \\ \hline

%Eigen \textit{et al.} (b rgbd+normals) \cite{Eigen:etal:ICCV2015} & 57.7 & \textbf{39.7} & 77.6 & \textbf{71.1} & 95.9 & \textbf{64.1} & \textbf{54.9} &\textbf{49.4} & 45.8 &\textbf{45.0}& 25.2 & 87.9 & \textbf{57.6}  \\ \hline
%55.1 & \textbf{52.1} & 45.8 & \textbf{45.0} & \textbf{41.9} & 88.7 & \textbf{57.7} \\ \hline
Hermans \textit{et al.}(rgbd+crf)\cite{Hermans:etal:ICRA2014} & 68.4 & N/A & \textbf{83.4} & 41.9 & 91.5 & 37.1 & 8.6 & N/A & 28.5 & 27.7 & 38.4 & 71.8 & 46.1 \\ \hline
%SceneNet-FT-NYU-DHA(1.5k) & 54.19& 62.38 & 38.18 & 73.31 & 47.02 & 37.57 & 43.60 & 27.56 & 16.71 & 71.16 & 17.43 \\ \hline
\end{tabular}
\vspace{0.5mm} \vspace{0.5mm}
\caption{Results on NYUv2 test data for 13 semantic classes. We see a similar pattern here --- adding synthetic data helps immensely in improving the performance of nearly all functional categories of objects using DHA as input channels. As expected, accuracy on \textit{books}, \textit{painting}, \textit{tv}, and \textit{windows}, is compromised highlighting that the role of depth as a modality to segment these objects is limited. Note that we recomputed the accuracies of \cite{Eigen:etal:ICCV2015} using their publicly available annotations of 320$\times$240 and resizing them to 224$\times$224. Hermans \textit{et al.} \cite{Hermans:etal:ICRA2014} use ``\textit{Decoration}" and ``\textit{Bookshelf}" instead of \textit{painting} and \textit{books} as the other two classes. Therefore, they are not directly comparable. Also, their annotations are not publicly available but we have still added their results in the table. Note that they use 640$\times$480. Poor performance of SceneNet-DHA and SceneNet-DO-DHA on \textit{tv} and \textit{windows} is mainly due to limited training data for these classes in SceneNet.} 
\label{table: CA breakdown for 13 classes}
\end{table*}

\paragraph{Results on 13 Class Semantic Segmentation: SUN RGB-D} We perform similar experiments on 13 classes on SUN RGB-D. The dataset provides 5,825 training images and 5,050 images for testing in total. This is one order of magnitude bigger in size than the NYUv2. As shown in Table \ref{table: comparison of various networks on GA and CA, SUNRGBD}, we observe similar trends \textit{i.e.} training on synthetic data and finetuning on real dataset helps in improving the accuracies by 1\% and 3.5\% in global and class accuracy respectively when comparing SUNRGD-DHA and SceneNet-FT-SUNRGBD-DHA. However, we quickly see diminshing returns when using dropout \textit{i.e.} SceneNet-FT-SUNRGBD-DHA and SceneNet-FT-SUNRGBD-DO-DHA perform nearly the same. Furthermore, SceneNet-FT-SUNRGBD-DO-DHA performs only 0.8\% and 0.9\% better in global and class accuracy respectively, compared to SUNRGBD-DO-DHA. It is worth remembering that when experimenting with NYUv2, the proportion of synthetic data was 10 times the real training data while a bigger SUN RGB-D dataset means that this proportion is only 2 times ($\frac{10,030}{5,825}\backsim 2$ ) the real training data, suggesting that further improvements can be possible either through another order of magnitude increase in data or a possible change in the architecture. Nonetheless, we set a benchmark on pure depth based segmentation on SUNRGBD dataset. Breakdown of class accuracies for different training variants is shown in Table \ref{table: CA breakdown for 13 classes, SUNRGBD}.

\subsection{Confidence in the final predictions}
\begin{figure}
\centerline{
\includegraphics[width=\linewidth]{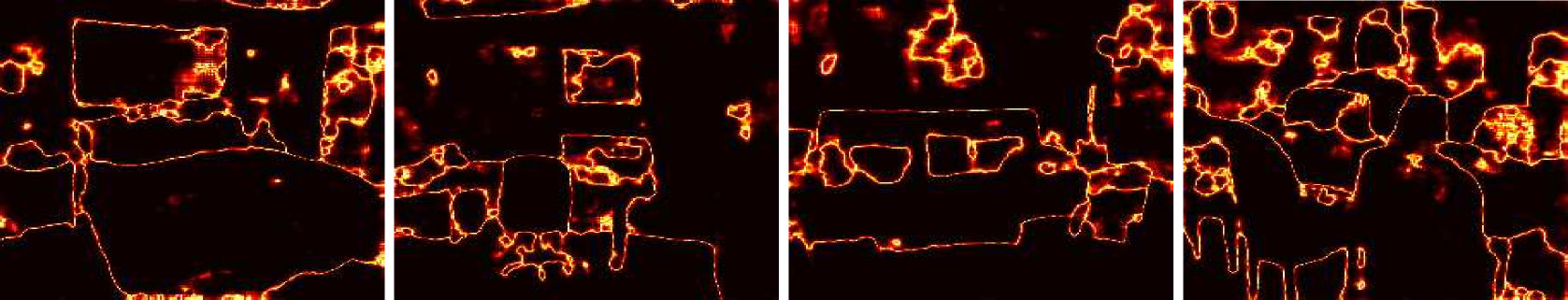}
}
\caption{\small
Ratio of probabilities of second best label to best label. Brighter colours represent higher uncertainty in the final prediction of SceneNet-FT-NYU-DHA for the images shown in Fig. \ref{fig:NYUv2results}.}
\label{fig:best_prob and div second best by best}
\end{figure}

We also plot the confidence in the final per-pixel segmentation label predictions as the ratio of probabilities of second best label to the best label. As seen clearly in the Fig. \ref{fig:best_prob and div second best by best}, higher uncertainty occurs mostly at the object boundaries. We also tried variation ratio of the final class labels returned by using dropout for each pixel at test time\cite{Gal:etal:arXiv2015} \cite{Kendall:etal:ARXIV2015} but did not find it any more informative.

\section{Conclusions}
We presented an effective solution to the problem of indoor scene understanding --- system trained with large number of rendered synthetic depth frames is able to achieve near state-of-the-art performance on per-pixel image labelling despite using only depth data. We specifically show that synthetic data offers a promising route in further improvements in the state-of-the-art and also introduce a new dataset of annotated 3D scenes to generate virtually unlimited training data. In future this dataset can be used to generate annotated videos to open up exciting possibilities in training networks on sequential data \textit{e.g.} RNNs, reinforcement learning, physical scene understanding and beyond. We hope to continue indoor scene segmentation on real world scenes with a reconstruction system running in the loop to bring real-time semantic segmentation using fused depth-maps.

\begin{table}
\centering
\begin{tabular}{l}
\textbf{13 Class Semantic Segmentation: NYUv2} \\ 
\end{tabular}
\begin{tabular}{|l|p{1.2cm}|p{1.2cm}|}
\hline
Training & global acc. & class acc. \\ \hline
Hermans \textit{et al.}(rgbd+crf) \cite{Hermans:etal:ICRA2014} & 54.2 & 48.0 \\ \hline
SceneNet-DHA & 54.4 & 37.1 \\ \hline
NYU-DHA & 63.6 & 47.2 \\ \hline
SceneNet-FT-NYU-DHA & 66.5 & 51.7 \\ \hline
SceneNet-DO-DHA & 55.3 & 37.6 \\ \hline
NYU-DO-DHA & 65.0 & 48.6 \\ \hline
SceneNet-FT-NYU-DO-DHA & 67.2 & 52.5 \\ \hline
Eigen \textit{et al.}(rgbd+normals) \cite{Eigen:etal:ICCV2015} & \textbf{68.0} & \textbf{60.8} \\ \hline
\end{tabular}\vspace{0.7mm}
\caption{The performance jump from NYU-DHA to SceneNet-FT-NYU-DHA follows similar trend --- the role of synthetic data is evident. Adding dropout shows only a marginal improvement. Overall increase in performance from NYU-DO-DHA to SceneNet-FT-NYU-DO-DHA is by 2.2\% and 3.9\% in global and class accuracy respectively. Note that we recomputed the accuracies of \cite{Eigen:etal:ICCV2015} using their publicly available annotations of 320$\times$240 and resizing them to 224$\times$224. Hermans \textit{et al.}\cite{Hermans:etal:ICRA2014} annotations are not publicly available but we have still added their results in the table. Moreover, they use 640$\times$480.}
\label{table: comparison of various networks on GA and CA for 13 classes}
\end{table}

\begin{table*}
\begin{tabular}{l}
\textbf{13 Class Semantic Segmentation: SUNRGBD} \\ 
\end{tabular}
\centering
\begin{tabular}{|l|p{0.5cm}|p{0.5cm}|p{0.5cm}|p{0.5cm}|p{0.5cm}|p{0.5cm}|p{0.5cm}|p{0.5cm}|p{0.5cm}|p{0.5cm}|p{0.5cm}|p{0.5cm}|p{0.5cm}|p{0.5cm}|}
\hline
Training  & \cellcolor{bedColor}\rotatebox{90}{bed} & \cellcolor{booksColor}\rotatebox{90}{books}  & \cellcolor{ceilColor}\rotatebox{90}{ceil.} & \cellcolor{chairColor}\rotatebox{90}{chair}  & \cellcolor{floorColor}\rotatebox{90}{floor}  & \cellcolor{furnColor}\rotatebox{90}{furn}   & \cellcolor{objsColor}\rotatebox{90}{objs.} & \cellcolor{paintColor}\rotatebox{90}{paint.} & \cellcolor{sofaColor}\rotatebox{90}{sofa}   & \cellcolor{tableColor}\rotatebox{90}{table}  & \cellcolor{tvColor}\rotatebox{90}{tv}     & \cellcolor{wallColor}\rotatebox{90}{wall}   & \cellcolor{windColor}\rotatebox{90}{window} \\ \hline
SceneNet-DHA & 33.2 & 2.5 & 40.6 & 54.0 & 71.1 & 26.2 & 22.1 & 9.5 & 15.0 & 29.2 & 0 & 89.2 & 0.0 \\ \hline
SceneNet-DO-DHA & 46.1 & 5.2 & 43.6 & 54.8 & 63.1 & 37.4 & 23.2 &  10.7 & 12.2 & 29.8 & 0 & 83.6 & 1.0 \\ \hline
SUNRGBD-DHA & 70.4 & 11.2 & 64.7 & 69.2 & \textbf{94.0} & 48.4 & 35.3 & 13.7 & 48.2 & 63.0 & 3.5 & \textbf{89.7} & 27.9 \\ \hline
SUNRGBD-DO-DHA & 73.6 & 16.6 & \textbf{71.6} & 70.1 & 93.5 & 47.9 & 38.7 & \textbf{17.2} & \textbf{58.5} & 61.8 & 6.8 & 88.7 & 33.9 \\ \hline
SceneNet-FT-SUNRGBD-DHA & 69.0 & \textbf{20.0} & 70.3 & 70.7 & 93.7 & 49.7 & 35.5 & 15.7 & 57.8 & \textbf{65.9} & \textbf{14.1}& 89.0 & 33.8 \\ \hline
SceneNet-FT-SUNRGBD-DO-DHA & \textbf{75.6} & 13.5 & 69.2 & \textbf{73.6} & 93.8 & \textbf{52.0} & \textbf{37.1} & 16.8 & 57.2 & 62.7 & 9.5 & 88.8 & \textbf{36.5} \\ \hline
\end{tabular}
\vspace{0.5mm} \vspace{0.5mm}
\caption{Results on SUNRGBD test data for 13 semantic classes. We see a similar pattern here --- adding synthetic data helps immensely in improving the performance of nearly all functional categories of objects using DHA as input channels. As expected, accuracy on \textit{books}, \textit{painting}, \textit{tv}, and \textit{windows}, is compromised highlighting that the role of depth as a modality to segment these objects is limited. } 
\label{table: CA breakdown for 13 classes, SUNRGBD}
\end{table*}

\begin{table}
\centering
\begin{tabular}{l}
\textbf{13 Class Semantic Segmentation: SUNRGBD} \\ 
\end{tabular}
\begin{tabular}{|l|p{0.9cm}|p{0.9cm}|}
\hline
Training & global acc. & class acc. \\ \hline
%Hermans \textit{et al.}(rgbd+crf) \cite{Hermans:etal:ICRA2014} & -- & 49.4 \\ \hline
SceneNet-DHA & 56.9 & 30.2 \\ \hline
SUNRGBD-DHA & 73.7 & 49.2 \\ \hline
SceneNet-FT-SUNRGBD-DHA & 74.7 & 52.7 \\ \hline
SceneNet-DO-DHA & 54.7 & 31.6 \\ \hline
SUNRGBD-DO-DHA & 74.2 & 52.2 \\ \hline
SceneNet-FT-SUNRGBD-DO-DHA & \textbf{75.0} & \textbf{53.1} \\ \hline
%Eigen \textit{et al.}(rgbd+normals) \cite{Eigen:etal:ICCV2015} & \textbf{87.4} & \textbf{63.0} \\ \hline
\end{tabular}\vspace{0.7mm}
\caption{Global and class accuracies for 13 class experiments on SUN RGB-D. We see improvements of 1\% and 3.5\% in global and class accuacy comparing SUNRGB-DHA and SceneNet-FT-SUNRGBD-DHA. However, when using dropout, SUNRGBD-DO-DHA, SceneNet-FT-SUNRGBD-DHA and SceneNet-FT-SUNRGBD-DO-DHA perform nearly the same suggesting that increase in data is only helpful up to a point and that further improvements can be possible either through another order of magnitude of data as seen in NYUv2 experiments or a possible change in the architecture. } 
\label{table: comparison of various networks on GA and CA, SUNRGBD}
\end{table}

{\small
\bibliographystyle{ieee}
\bibliography{egbib}
}
%
%%\bibliographystyle{ieeetr}
%%\bibliography{robotvision.bib}

\end{document}